\documentclass[11pt]{article}

\usepackage[final]{acl}

\usepackage{times}
\usepackage{latexsym}
\usepackage{float}
\usepackage[T1]{fontenc}
\usepackage{amsmath}
\usepackage{amssymb}
\usepackage{subcaption}

\usepackage[utf8]{inputenc}
\usepackage{booktabs}
\usepackage{microtype}

\usepackage{inconsolata}

\usepackage{graphicx}
\usepackage{fvextra}
\usepackage{fontawesome6}
\usepackage{multirow}
\usepackage{tabularx}
\usepackage{array}
\usepackage{simpleicons}

\usepackage{xcolor}
\definecolor{githubcolor}{HTML}{181717}
\definecolor{hfcolor}{HTML}{FFD21E}

\usepackage{titlesec}

\titleclass{\subsubsubsection}{straight}[\subsubsection]
\newcounter{subsubsubsection}[subsubsection]
\renewcommand{\thesubsubsubsection}{\thesubsubsection.\arabic{subsubsubsection}}

\titleformat{\subsubsubsection}
  {\normalfont\normalsize\bfseries}{\thesubsubsubsection}{1em}{}
\titlespacing*{\subsubsubsection}
  {0pt}{3.25ex plus 1ex minus .2ex}{1.5ex plus .2ex}

\makeatletter
\def\toclevel@subsubsubsection{4}
\def\l@subsubsubsection{\@dottedtocline{4}{7.0em}{4.1em}}
\makeatother
\title{\texttt{PlurVA-LLM-2026 Shared Task Track-1:} Pluralistic Value Alignment in LLMs via Multilingual Fine-Tuning and Threshold Calibration}

\author{Vihindi Kotalawala$^\heartsuit$ \& Nevidu Jayatilleke$^\diamondsuit$ \\
$^\heartsuit$School of Computing, Informatics Institute of Technology, Sri Lanka\\
$^\diamondsuit$Department of Computer Science \& Engineering, 
University of Moratuwa, Sri Lanka \\
\texttt{vihindi.20220672@iit.ac.lk, nevidu.25@cse.mrt.ac.lk}
}

\begin{document}
\maketitle


\begin{abstract}
We present our system for the \textit{PlurVA-LLM 2026 Shared Task Track-1}, which focuses on pluralistic value alignment in the contexts of China, Indonesia, and Sri Lanka. For this resource-constrained track, we fine-tuned \texttt{Llama-3.1-8B-Instruct} using 4-bit QLoRA. Our approach combines option-permutation augmentation for Chinese data, annotator vote expansion for Indonesian data, and binary reformulation with SinhalaMMLU augmentation for Sri Lankan data. We further applied conditional threshold calibration to the predictions for the Sri Lankan data. The final system achieved accuracies of 0.785 for Chinese, 0.715 for Indonesian, and 0.916 for Sri Lankan, resulting in an overall macro-average accuracy of 0.805.
\vspace{-0.8cm}
\begin{center}
    \href{https://github.com/Vihindi/pluralva_shared_task}{\textcolor{githubcolor}{\faGithub}~Code}
    \quad
    \href{https://huggingface.co/Vihindi-K/PlurValLM-Llama-3.1-8B-ZH-ID-SI}{\raisebox{-0.35em}{\includegraphics[height=1.5em]{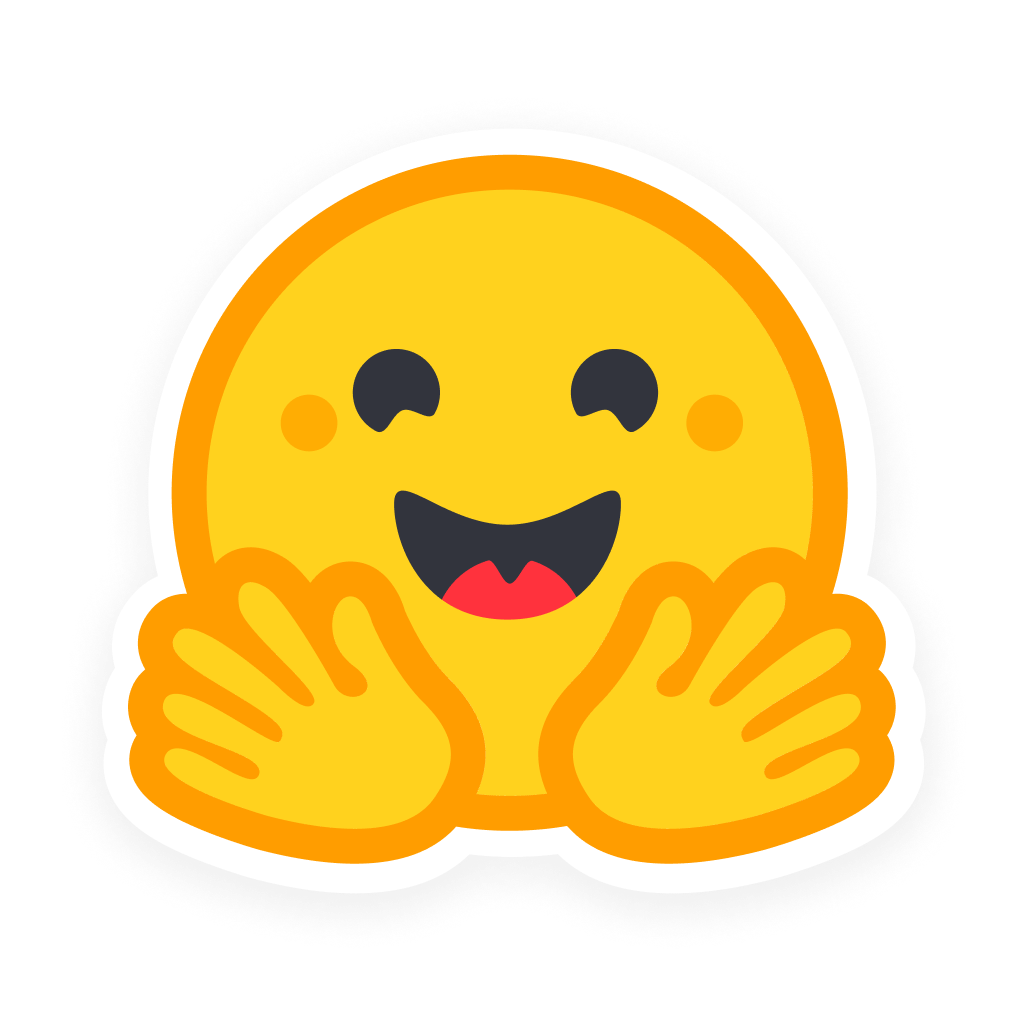}}~Model}
    \quad
    \href{https://huggingface.co/datasets/TJUNLP/PlurVA-LLM-ST-2026}{\raisebox{-0.35em}{\includegraphics[height=1.5em]{figures/hf-logo.png}}~Dataset}
\end{center}
\end{abstract}

\section{Introduction}
\label{sec:intro}

Large Language Models (LLMs) are increasingly used across diverse societies, yet conventional alignment often assumes a single set of human preferences. Human values vary across populations, motivating models that can represent diverse perspectives rather than a single universal preference \cite{sorensen2024roadmap}. Existing alignment methods may instead learn an averaged preference and struggle to represent differences across cultures and communities \cite{feng-etal-2024-modular}. Multilingual capability alone also does not guarantee consistent value alignment, particularly across high- and low-resource languages \cite{xu-etal-2024-exploring-multilingual}. 

The PlurVA-LLM 2026 Shared Task addresses this challenge by evaluating whether one system can make locally grounded value judgements across Chinese, Indonesian, and Sri Lankan contexts.
The task combines three datasets with distinct cultural foundations and annotation structures. \texttt{D2VBench} presents value conflicts through daily-life scenarios from China \cite{hao2026d2vbench}. \texttt{Pancasila-Dilemmas} evaluates Indonesian judgements grounded in Pancasila and provides five citizen annotations for each question \cite{martin2026pancasila}. \texttt{LKValues} evaluates societal values derived from Sri Lankan participants and expressed through Sinhala language \cite{muthugala2026lkvalues}.

We adopt a shared multilingual fine-tuning approach while adapting supervision to each dataset's characteristics. A shared adapter enables the model to learn across multiple cultural contexts, while country-specific prompts indicate the relevant value setting for each example \cite{xu-etal-2025-self}. At the same time, dataset-specific preprocessing is used to account for differences in annotation formats, label structures, and observed preference distributions across the three datasets \cite{pezeshkpour-hruschka-2024-large,sorensen2024roadmap}. This allows us to retain dataset-specific value signals without requiring a completely separate model for each cultural context. For Sri Lankan data, where the supervision is comparatively limited and imbalanced, we further apply conditional threshold calibration at inference time to better map the model's statement-level confidence into the final four-way prediction \cite{esposito2021ghost}. Overall, our approach combines learning across all three datasets with specific adjustments for the distinct challenges present in each.

\section{Related Work}

Prior work on value alignment has increasingly considered pluralistic, multilingual, and culturally grounded preferences. Relevant studies span methods for representing heterogeneous human values, adapting models across cultural and linguistic contexts, and addressing challenges arising from preference disagreement, limited low-resource data, and cross-lingual variation. We review these directions below in relation to our multilingual value-alignment setting.

\subsection{Pluralistic Value Alignment}

Pluralistic value alignment challenges the assumption that an LLM should optimise toward a single aggregate human preference. But as discussed in Section~\ref{sec:intro}, relying mostly on a single reward signal can fail to represent diverse human preferences \cite{chakraborty2024maxmin}, while diversified annotator preferences can also reduce the effectiveness of standard alignment methods \cite{zeng-etal-2024-diversified}. There are three objectives for pluralistic alignment: Overton, steerable, and distributional pluralism, with distributional pluralism aiming to reflect the preference distribution of a community \cite{sorensen2024roadmap}. Several approaches have consequently attempted to model such diversity explicitly. MaxMin-RLHF learns a mixture of reward models and optimises for diverse preference groups \cite{chakraborty2024maxmin}, while Modular Pluralism combines a base LLM with specialised community models to support different value perspectives \cite{feng-etal-2024-modular}.

Value pluralism also extends across languages and societal contexts, where models may need to represent different locally grounded values rather than transfer a single dominant value system across populations. Existing LLMs can reflect the values of particular groups by default, motivating the development of methods specifically designed for cultural-value adaptation \cite{liu-etal-2025-cultural}. CultureSPA addresses this through culture-joint and culture-specific fine-tuning, enabling a single model to serve multiple sets of culturally situated human values \cite{xu-etal-2025-self}. Evaluation work has similarly moved beyond generic preference benchmarks. WorldValuesBench uses demographic contexts and real-world survey responses to evaluate whether language models reproduce diverse human-value distributions, and shows that this remains challenging even for strong models \cite{zhao-etal-2024-worldvaluesbench}.

\subsection{Challenges in Value Alignment}

Translating these pluralistic objectives into effective model training introduces several additional challenges. There is insufficient human preference data in many languages, making it harder to align models with the preferences expressed in those languages \cite{yang-etal-2025-implicit}, while human-value representations can exhibit cross-lingual inconsistencies and asymmetric transfer between high- and low-resource languages \cite{xu-etal-2024-exploring-multilingual}. Recent Sri Lankan value-alignment work likewise reports persistent low-resource and cultural value-alignment gaps, with fine-tuning gains varying across model families \cite{muthugala2026lkvalues}. The structure of the supervision itself is also important. Preference datasets constructed using rigid annotation or aggregation schemes can suppress underlying preference heterogeneity \cite{chen2024pal}. Standard preference-learning methods may not clearly distinguish between full agreement and a majority preference when annotators disagree \cite{zhang2024diverging}.
Consequently, collapsing multiple annotations into a single majority label can remove meaningful minority judgments when disagreement reflects genuine value differences rather than annotation noise. Fine-tuning with such heterogeneous preferences is further affected by both model capacity and available data \cite{zeng-etal-2024-diversified}. In multiple-choice value tasks, answer-option order can introduce an additional source of bias \cite{pezeshkpour-hruschka-2024-large}.

Prior work has explored pluralistic preference learning, multilingual value alignment, and culturally grounded evaluation, often highlighting performance differences between high-resource and low-resource settings. In contrast, our setting requires a single model to perform well across all three datasets despite differences in language resources, annotation structures, and data availability. The goal is not only to measure these disparities but also to maximise performance across both high- and low-resource settings while avoiding improvements on one dataset at the expense of another. We address this by combining shared multilingual learning with dataset-specific supervision and inference strategies, aiming for consistently strong performance across the different value contexts and data conditions.

\subsection{Language and Resource Context}

The three communities involved in this shared task, Chinese, Indonesian, and Sri Lankan, are represented by the languages Chinese, Indonesian, and Sinhala. Each language corresponds to varying levels of data resources. Consequently, the strategy employed for each language is tailored to the availability of linguistic resources.

\paragraph{Chinese:}
Standard Chinese (Mandarin) is the official and predominant language of China and belongs to the Sino-Tibetan language family~\cite{yang-etal-2022-cino}. It is written using Chinese characters and is classified as a Category~5 (The Winners) language, representing a highly resourced NLP setting~\cite{joshi-etal-2020-state,ranathunga-de-silva-2022-languages}. Among the three languages in this work, Chinese provides the highest resource capacity.

\paragraph{Indonesian:}
Indonesian (Bahasa Indonesia) is the national language of Indonesia and is widely used across its multilingual population~\cite{gunarso-riza-2016-overview}. It has complex word-formation patterns and is spoken by more than 250 million people~\cite{kamajaya-moeljadi-2025-indomorph}. \citet{ranathunga-de-silva-2022-languages} classify Indonesian as Category~4 (The Underdogs). 

\paragraph{Sinhala:}
Sinhala is an Indo-Aryan language spoken primarily in Sri Lanka by approximately 16 million first-language speakers~\cite{jayatilleke-de-silva-2025-sidiac}. It uses a Brahmi-derived writing system~\cite{de2019survey} and is classified as a lower-resourced Category~2 (The Hopefuls) language~\cite{ranathunga-de-silva-2022-languages}. Therefore, Sinhala is the most resource-constrained of the three languages considered in this work.

\section{Methodology}

The three PlurVA-LLM datasets differ substantially in their annotation and output structures. Chinese is a four-option multiple-choice task with one gold answer. Indonesian also contains four options, but each question is annotated by five individuals and may contain disagreement. Sri Lankan contains two candidate statements and requires one of four final labels: only statement A is correct, only statement B is correct, both are correct, or neither is correct.

We developed country-specific input and target transformations while retaining a single shared model architecture. All examples were converted into chat-formatted supervised fine-tuning records. The final system used  \texttt{Meta Llama-3.1-8B-Instruct} as its base model and was adapted using 4-bit Quantised Low-Rank Adaptation (QLoRA) \cite{dettmers2023qlora}. A single shared LoRA adapter was trained jointly on all 3 datasets: Chinese, Indonesian, and Sri Lankan.

\begin{figure*}
    \centering
    \includegraphics[width=\textwidth]{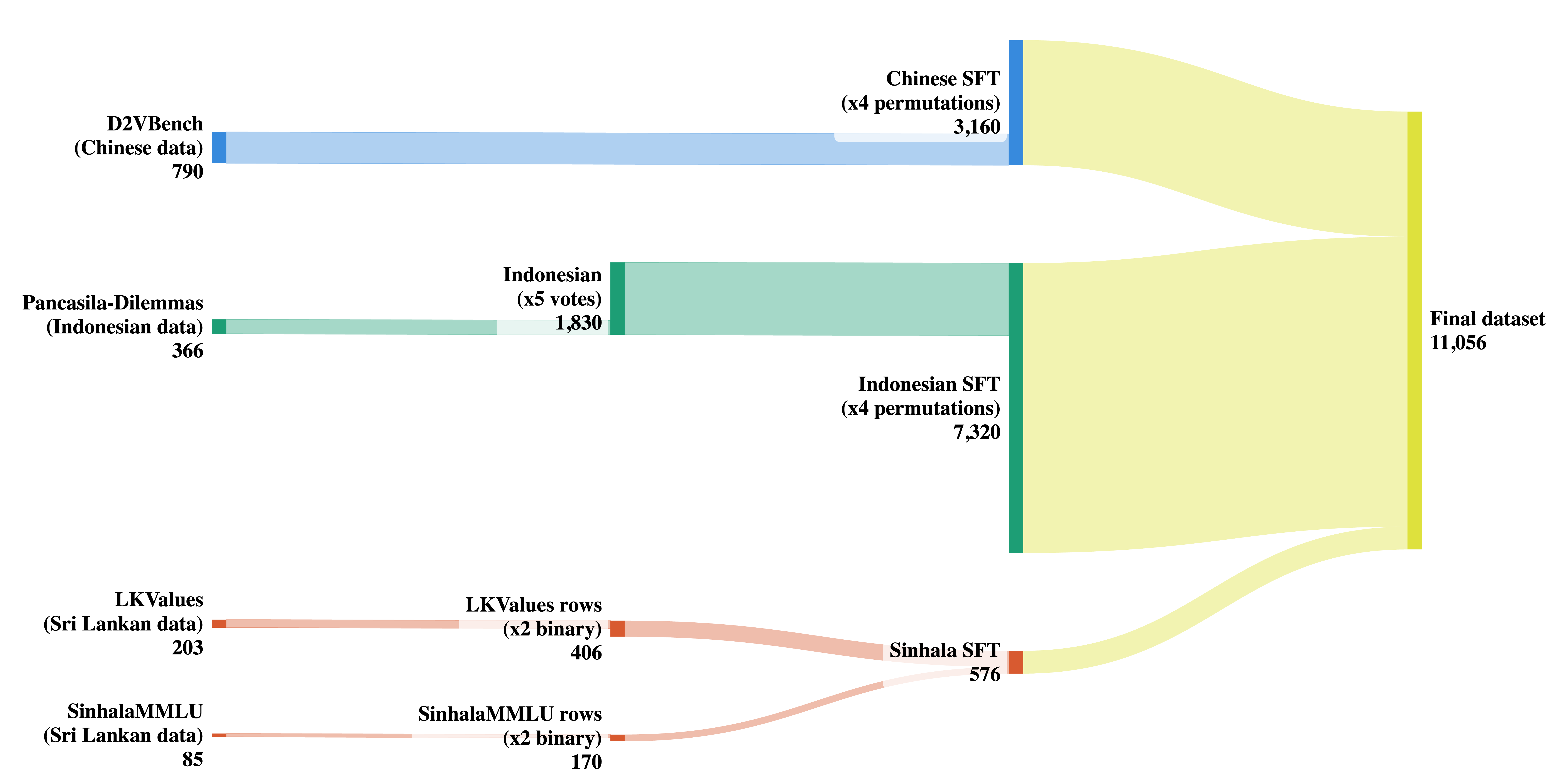}
    \caption{Data preparation and augmentation pipeline for the PlurVA-LLM training set. Flow widths are proportional to the number of training records produced at each stage for \texttt{D2VBench}, \texttt{Pancasila-Dilemmas}, \texttt{LKValues}, and \texttt{SinhalaMMLU} before merging into the final multilingual SFT dataset.}
    \label{fig:data_pipeline}
\end{figure*}

All available labelled records were used for final training; no validation partition was reserved since the dataset was extremely small. The final training set contained 11,056 supervised records after country-specific transformations and augmentation shown in Figure~\ref{fig:data_pipeline}, while the total evaluation set was 5475 (Chinese: 3,210, Indonesian:1,468, Sri Lankan:797 ).

\subsection{Base-model Selection}

We first compared five instruction-tuned models under zero-shot prompting. Table~\ref{tab:zero-shot} reports the macro average and country-level accuracies.

Although the strongest individual-language results varied across models. \texttt{Llama-3.1-8B-Instruct} for Chinese at 0.5, \texttt{Aya Expanse 8B} for Indonesian at 0.7027, and \texttt{Gemma 3n-E4B-it} for Sinhala at 0.8293 show the best-performing models for each language. But \texttt{Llama-3.1-8B-Instruct} achieved the highest macro-average of 0.6394, surpassing other models. It also provided a comparatively balanced performance across the three languages. Therefore, we selected it as the shared base model.

\begin{table}[t]
\centering
\small
\resizebox{\columnwidth}{!}{%
\begin{tabular}{lcccc}
\hline
\textbf{Model} & \textbf{Chinese} & \textbf{Indonesian} & \textbf{Sinhala} & \textbf{Macro} \\
\hline
\texttt{Qwen2.5-7B-Instruct}       & 0.4684 & 0.6351 & 0.4146 & 0.5060 \\
\texttt{Aya Expanse 8B}            & 0.4494 & \textbf{0.7027} & 0.3415 & 0.4978 \\
\texttt{Llama-SEA-LION-v3-8B-IT}  & 0.4620 & 0.6757 & 0.6829 & 0.6069 \\
\texttt{Llama-3.1-8B-Instruct}     & \textbf{0.5000} & 0.6622 & 0.7561 & \textbf{0.6394} \\
\texttt{Gemma 3n-E4B-it}           & 0.3608 & 0.6757 & \textbf{0.8293} & 0.6219 \\
\hline
\end{tabular}
}%
\caption{Zero-shot model comparison.}
\label{tab:zero-shot}
\end{table}

\subsection{Country-specific Supervision }

\begin{table}[!t]
\centering
\renewcommand{\arraystretch}{1.25}
\resizebox{0.95\columnwidth}{!}{
\begin{tabular}{c|l|c}
\hline
\textbf{Language} &
\textbf{Experiment} &
\textbf{Score} \\
\hline
\multirow{2}{*}{Chinese}
& \textbf{$-$} Option-permutation ensembling
& 0.779 \\
& \textbf{+} Option-permutation ensembling
& \textbf{0.785} \\
\hline
\multirow{3}{*}{Indonesian}
& Majority-vote hard-label training
& 0.651 \\
\cline{2-3}
& Annotator-vote expansion
& 0.689 \\
& \hspace*{1.2em}\textbf{+} Option permutation
& \textbf{0.715} \\
\hline
\multirow{5}{*}{Sri Lankan}
& 4-way classification (w/ adapter)\textsuperscript{\textdagger}
& 0.832 \\
& \hspace*{1.2em}\textbf{+} Binary formulation (w/ adapter)
& \textbf{0.834} \\
\cline{2-3}
& Binary formulation (w/o adapter)\textsuperscript{\textasteriskcentered}
& 0.873 \\
& \hspace*{1.2em}\textbf{+} MMLU augmentation
& \textbf{0.890} \\
& \hspace*{2.4em}\textbf{+} Conditional calibration
& \textbf{0.916} \\
\hline
\end{tabular}}
\vskip 4pt
\begin{minipage}{0.95\columnwidth}
  \centering\scriptsize
  \textdagger~Language-specific adapter.\quad
  \textasteriskcentered~Multilingual shared adapter.
\end{minipage}
\caption{Language-level experiments and selected configurations.}
\label{tab:language_ablation}
\end{table}

\subsubsection{Chinese}

The Chinese training set contained 790 questions. Each question included a value label, a question written in Chinese, four answer options, and one gold label from $\{A,B,C,D\}$. The original label distribution was imbalanced: 237 A, 123 B, 136 C, and 294 D.

The prompt instructions, the question, and the four answer options were written in Chinese. The model was prompted to return exactly one answer line in the form \texttt{Answer: X} as depicted in Figure~\ref{fig:chinese-main-prompt} in  Appendix~\ref{app:prompt-templates}.

To reduce dependence on answer position, we applied cyclic option-permutation augmentation. For each question, four orderings of the options were constructed. When the options were shifted, the gold label was remapped to continue identifying the same answer content. As shown in Table~\ref{tab:language_ablation}, incorporating option permutation into the Chinese pipeline improved accuracy from 0.779 to 0.785.

The four cyclic orderings produced 3,160 Chinese SFT rows:
$790 \times 4 = 3,160$. Because every original answer appeared once in each displayed position across the four cyclic permutations, the augmented target distribution contained exactly 790 instances of each output label. This simultaneously increased the amount of supervision and removed the original output-letter imbalance.

At inference, the same four cyclic orderings were scored. Each of the 3,210 original records was expanded into four permutations, resulting in a total of 12,840 records for prediction. Probabilities were mapped back to the original option identities and averaged before selecting the final answer. This made the final prediction less sensitive to the order in which options were displayed.

\subsubsection{Indonesian}

The Indonesian set contained 366 questions. Each question contained four answer options and five annotator votes, resulting in 1,830 individual annotations.  The raw vote totals were 505 A, 402 B, 544 C, and 379 D. Of the 366 questions, 294 had a single consensus answer, while 72 had two answers in their consensus set.
A majority-vote transformation would discard this disagreement and force each question to have a single hard target. In our experiments, as shown in Table~\ref{tab:language_ablation}, the majority-vote approach achieved an Indonesian accuracy of 0.651. Therefore, we retained all five annotations through vote expansion, thereby improving accuracy to 0.689.

The prompt instructions, the question, and the four options were written in Indonesian. The model was prompted to return only the answer in the form \texttt{Answer: X} as shown in Figure~\ref{fig:indonesian-main-prompt} in Appendix~\ref{app:prompt-templates}.

For a question with annotator labels $y_1,\ldots,y_5$, we created one training record for every vote. 
For each Indonesian question, the five annotator votes were retained as separate training targets. Their contributions were represented through the log-probabilities of the corresponding answers, so answers receiving more votes contributed proportionally more to the learning objective while minority judgments were still preserved. This allowed the empirical annotator distribution to be learned without introducing a custom soft-label loss.

We additionally applied the same four cyclic option permutations used for Chinese. Each Indonesian question consequently generated $(5\ \text{votes} \times 4\ \text{permutations} = 20)$ training records, producing 7,320 Indonesian SFT rows, which increased the accuracy from 0.689 to 0.715, as shown in Table~\ref{tab:language_ablation}.

During inference, we employed the same strategy used for the Chinese evaluation dataset. Each of the 1,468 original records was expanded into four permutations, resulting in a total of 5,872 records for prediction. The predicted probabilities across the four permutations were then averaged to determine the final answer for each original record.

\subsubsection{Sri Lankan}

The labelled Sri Lankan training set contained 203 questions written entirely in Sinhala. Its original four-way label distribution was 107 A, 33 B, 43 Both, and 20 neither-correct. This direct four-way training would expose the model to a considerable output-class imbalance.

Since the two statements were independent in the dataset, we decomposed every question into two independent binary decisions:
\[
\begin{gathered}
z_A = \mathbb{I}[\text{statement A is correct}] \\
z_B = \mathbb{I}[\text{statement B is correct}]
\end{gathered}
\]

Each question generated two SFT records: one presenting statement A and one presenting statement B. The target for each record was either \texttt{Answer: Yes} or \texttt{Answer: No}. At inference, the two decisions were recombined as follows:

\[
(z_A,z_B)=
\begin{cases}
(1,0) &\rightarrow A,\\
(0,1) &\rightarrow B,\\
(1,1) &\rightarrow \mathrm{Both},\\
(0,0) &\rightarrow 0.
\end{cases}
\]

The 203 original questions produced 406 binary records, consisting of 226 Yes targets and 180 No targets. This reformulation reduced the severity of the original four-class imbalance while preserving all four possible final outcomes. During the multi-adapter experiment, in which separate language-specific adapters were trained, we compared the four-way and binary formulations for Sinhala. As shown in Table~\ref{tab:language_ablation}, the binary formulation performed slightly better, improving accuracy from 0.832 to 0.834.

The prompt retained the original Sinhala question and candidate statement. The system and task instructions were written in English and described the broader Sri Lankan context, including the Sinhalese, Tamil, Muslim, and Burgher communities. The corresponding Sri Lankan value label was also included. The model was prompted to return only \texttt{Answer: Yes} or \texttt{Answer: No} as shown in Figure~\ref{fig:sinhala-binary-main-prompt} in Appendix~\ref{app:prompt-templates}.

During inference, each record was split into two separate questions, following the same approach used during training. As a result, the original 797 records were expanded into 1,594 records for prediction.

\subsubsubsection{SinhalaMMLU Augmentation}

The Sri Lankan dataset had substantially fewer training examples than the Chinese and Indonesian datasets. We augmented it with manually selected questions from \texttt{SinhalaMMLU} \cite{pramodya-etal-2025-sinhalammlu}. From the original resource, we selected 68 Civics questions and 17 Health and Physical Science questions, giving 85 additional questions.

Selection was performed manually according to relevance to the shared task. Each selected MMLU question originally contained four answer options and one correct option. We converted these records into the same two-statement Sinhala format used by the main dataset.

\begin{figure}
    \centering
    \includegraphics[width=1.0\linewidth]{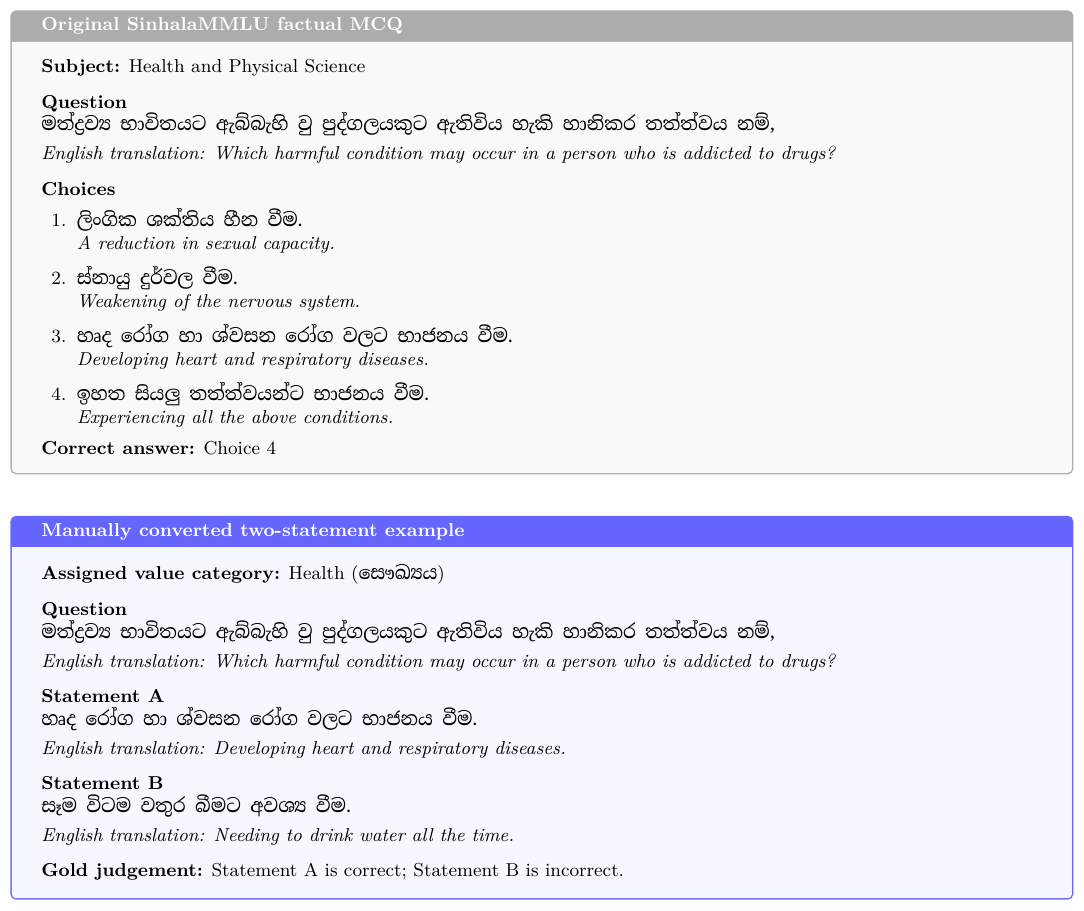}
    \caption{Example of manually adapting the answers of a
SinhalaMMLU Health and Physical Science question into two
candidate statements.}
    \label{fig:mmlu_health_conversion}
\end{figure}

Two conversion strategies were used:

\begin{enumerate}
    \item When the original question could be retained, the correct answer and the most plausible alternative among the remaining answers were selected as the two candidate statements.
    \item When a meaningful two-statement record could not be produced directly, the question and candidate statements were manually amended or rewritten while preserving the relevant knowledge and value judgment by the authors of this study, who are native Sinhala speakers. Figure~\ref{fig:mmlu_health_conversion} presents an example in
which the answer choices of a SinhalaMMLU Health and Physical
Science question were manually adapted into two candidate
statements.
\end{enumerate}

Each converted question contained one correct and one incorrect statement, producing 85 Yes and 85 No records. The MMLU augmentation\footnote{\href{https://huggingface.co/datasets/Vihindi-K/plurva-llm-sinhala-augmented-dataset}{\raisebox{-0.35em}{\includegraphics[height=1.5em]{figures/hf-logo.png}}~PlurVA-LLM-Sinhala-Augmented}} contributed 170 balanced binary records.

\texttt{Gemini 3.1 Flash} was used only to assign one value-category label to each converted record. All generated value labels were manually refined by the authors, who are native Sinhala speakers, before inclusion.

Combining the main Sinhala data with MMLU augmentation yielded 288 questions and 576 binary SFT rows: $(203+85)\times2=576.$
The combined binary distribution contained 311 Yes and 265 No targets.
As shown in Table~\ref{tab:language_ablation}, this augmentation improved accuracy from 0.873 to 0.890.

\subsection{Prompt and Target Design}

All three countries' prompts explicitly included the supplied value category. Chinese and Indonesian used instructions in their respective languages, while Sri Lankan retained the original Sinhala content with concise English task instructions. Prompts for the respective langagues are mentioned in Appendix~\ref{app:prompt-templates}.

During development, we additionally experimented with rationale supervision and value-description summaries. These configurations and their corresponding performance are reported in Appendix~\ref{app:rationale-value-summary}. However, neither rationales nor value summaries were included in the final selected model because they did not improve the overall performance.

This result suggests that the extra explanations added unnecessary or noisy information for the classification task. \citet{zhu-etal-2025-rationales} has similarly shown that adding rationales does not always improve language-model performance and can sometimes reduce accuracy. More generally, adding extra training signals can lead to negative transfer when they introduce task-irrelevant information or misleading patterns. Since the required outputs in this task were short categorical decisions, directly supervising the target labels provided a more focused learning signal.

Every assistant target in the final model consisted only of a short machine-readable answer marker: ``\texttt{Answer: A/B/C/D}" for Chinese and Indonesian, or
``\texttt{Answer: Yes/No}" for Sri Lankan.

\subsection{QLoRA Fine-tuning}

The model was fine-tuned using QLoRA. Trainable LoRA matrices were inserted into seven attention and feed-forward projections:

\[
\begin{aligned}
\{&\texttt{q\_proj},\texttt{k\_proj},\texttt{v\_proj},\texttt{o\_proj},\\
  &\texttt{gate\_proj},\texttt{up\_proj},\texttt{down\_proj}\}.
\end{aligned}
\]

\begin{table}[t]
\centering
\small
\resizebox{\columnwidth}{!}{%
\begin{tabular}{ll}
\hline
\textbf{Hyperparameter} & \textbf{Final value} \\
\hline
Base model &  \texttt{Meta Llama-3.1-8B-Instruct} \\
Quantisation & 4-bit NF4 with double quantization \\
Compute data type & bfloat16 \\
Epochs & 2 \\
Peak learning rate & $1\times10^{-4}$ \\
Learning-rate schedule & Cosine decay \\
Warm-up ratio & 0.05 \\
Per-device batch size & 2 \\
Gradient accumulation & 10 \\
Effective batch size & 20 \\
Maximum sequence length & 1,536 \\
LoRA rank & 16 \\
LoRA alpha & 32 \\
LoRA dropout & 0.05 \\
Maximum gradient norm & 1.0 \\
Weight decay & 0.0 \\
Random seed & 42 \\
\hline
\end{tabular}
}%
\caption{Hyperparameters of the selected shared adapter.}
\label{tab:hyperparameters}
\end{table}

\begin{table*}[ht]
\centering
\resizebox{0.65\textwidth}{!}{%
\begin{tabular}{l|c|c|c|c}
\hline
\textbf{Adapter approach} & \textbf{Chinese} & \textbf{Indonesian} & \textbf{Sinhala} & \textbf{Macro avg} \\ \hline
Language-specific adapters & \textbf{0.788} & 0.687 & 0.834 & 0.770 \\\hline
Our approach (shared adapter) & 0.785 & 0.715 & 0.89 & 0.796 \\ 
\hspace*{1.2em}\textbf{+} threshold calibration & 0.785 & \textbf{0.715} & \textbf{0.916} & \textbf{0.805}\\ \hline

\end{tabular}
}%
\caption{Comparison of multiple language-specific adapters and a shared multilingual adapter}
\label{tab:adapter_comparison}
\end{table*}

Training used a standard causal language model cross-entropy with the prompt tokens masked, so the loss was computed only over the assistant's response. Since the targets contained only the final answer markers, the optimisation focused directly on the required output labels.

We compared separate language-specific adapters with our proposed shared multilingual adapter, jointly trained on all three languages. As shown in Table~\ref{tab:adapter_comparison}, the shared adapter achieved the highest macro-average accuracy of 0.796, compared with 0.770 for the language-specific setup, and was selected as our final approach. The shared setup was particularly effective for Indonesian and Sri Lankan data, reaching 0.715 and 0.89, respectively. This is consistent with~\citet{xu-etal-2025-self}, who report that joint fine-tuning can outperform culture-specific tuning by enabling knowledge sharing across groups. The final hyperparameters used for the best model are depicted in Table~\ref{tab:hyperparameters}.

\subsubsection{Conditional Threshold Calibration}

After fine-tuning, we calibrated the Sri Lankan decision thresholds using five-fold cross-validation on the training set. Because the original four-way prediction was reconstructed from two binary statement-level decisions, applying a fixed threshold of 0.5 to both statements did not provide the best class balance. \citet{esposito2021ghost} has similarly shown that fixed decision thresholds can be suboptimal under class imbalance and that post hoc threshold adjustment can improve classification performance. In particular, the model tended to over-predict cases in which statement B was considered valid. We used the five folds to identify suitable directions and ranges for adjusting the conditional thresholds.

The final prediction was determined using three thresholds. We first evaluated statement B using $\tau_B$. The threshold for statement A was then selected conditionally: $\tau_{A|B=0}$ was applied when B was predicted as invalid, while $\tau_{A|B=1}$ was applied when B was predicted as valid. The resulting binary decisions were then mapped back to the original four classes: A, B, Both, and 0.

\begin{figure}
    \centering
    \includegraphics[width=1.0\linewidth]{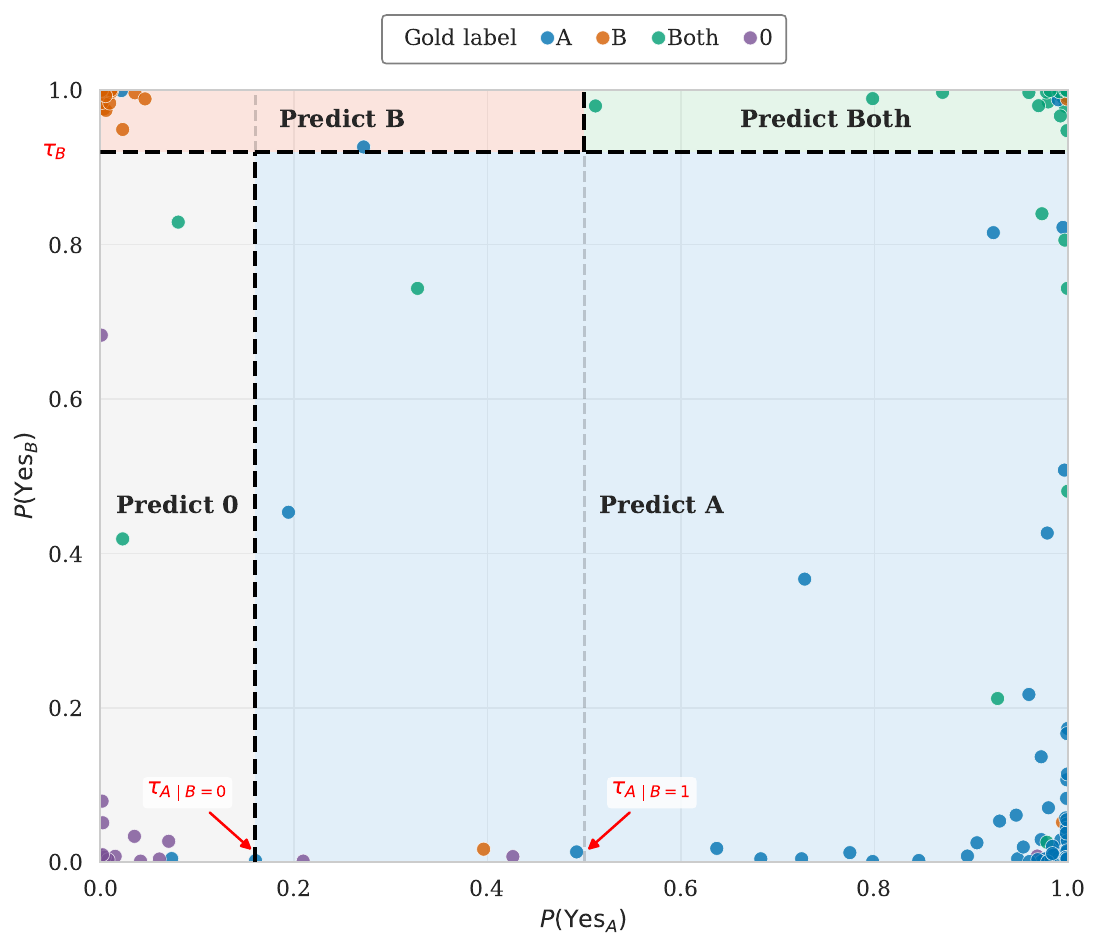}
    \caption{Conditional decision boundaries mapping Sinhala statement-level probabilities to the final classes A, B, Both, and 0.}
    \label{fig:decision_boundry}
\end{figure}

Figure~\ref{fig:decision_boundry} visualises the resulting decision regions. The horizontal boundary $\tau_B$ separates cases where B is predicted as valid from those where it is not. Below this boundary, $\tau_{A|B=0}$ separates predictions 0 and A, whereas above it, $\tau_{A|B=1}$ separates B and Both. Thus, the threshold applied to statement A depends on the preceding decision for statement B. A further detailed threshold sensitivity analysis is presented in Appendix~\ref{app:line_chart_app}.

Based on the five-fold cross-validation results, we narrowed the search to a smaller set of candidate threshold combinations. These configurations are reported in Table~\ref{tab:sinhala_calibration}, beginning with the default 0.5 thresholds and progressively adjusting $\tau_B$ and $\tau_{A|B=0}$ while keeping $\tau_{A|B=1}=0.50$.

\begin{table}[t]
\centering
\resizebox{0.8\columnwidth}{!}{%
\begin{tabular}{ccccc}
\hline
\textbf{Exp.} & $\boldsymbol{\tau_B}$ &
$\boldsymbol{\tau_{A|B=0}}$ &
$\boldsymbol{\tau_{A|B=1}}$ &
\textbf{Accuracy} \\
\hline
E0 & 0.50 & 0.50 & 0.50 & 0.8900 \\
E1 & 0.70 & 0.40 & 0.50 & 0.9100 \\
E2 & 0.80 & 0.30 & 0.50 & 0.9100 \\
E3 & 0.85 & 0.25 & 0.50 & 0.9100 \\
E4 & 0.90 & 0.20 & 0.50 & 0.9133 \\
E5 & 0.92 & 0.16 & 0.50 & \textbf{0.916} \\
\hline
\end{tabular}}
\caption{Conditional threshold calibration for the Sinhala statement-level predictions.}
\label{tab:sinhala_calibration}
\end{table}

The progression in Table~\ref{tab:sinhala_calibration} shows that increasing $\tau_B$ while decreasing $\tau_{A|B=0}$ improved performance over the default configuration. The selected setting used $\tau_B=0.92$, $\tau_{A|B=0}=0.16$, and $\tau_{A|B=1}=0.50$, improving accuracy from 0.8900 with the default thresholds to 0.916. As shown in Table~\ref{tab:adapter_comparison}, applying this calibration to Sri Lankan data with our shared multilingual adapter produced the final selected model, achieving the highest macro-average accuracy of 0.805.

\begin{figure}
    \centering
    \includegraphics[width=1.0\linewidth]{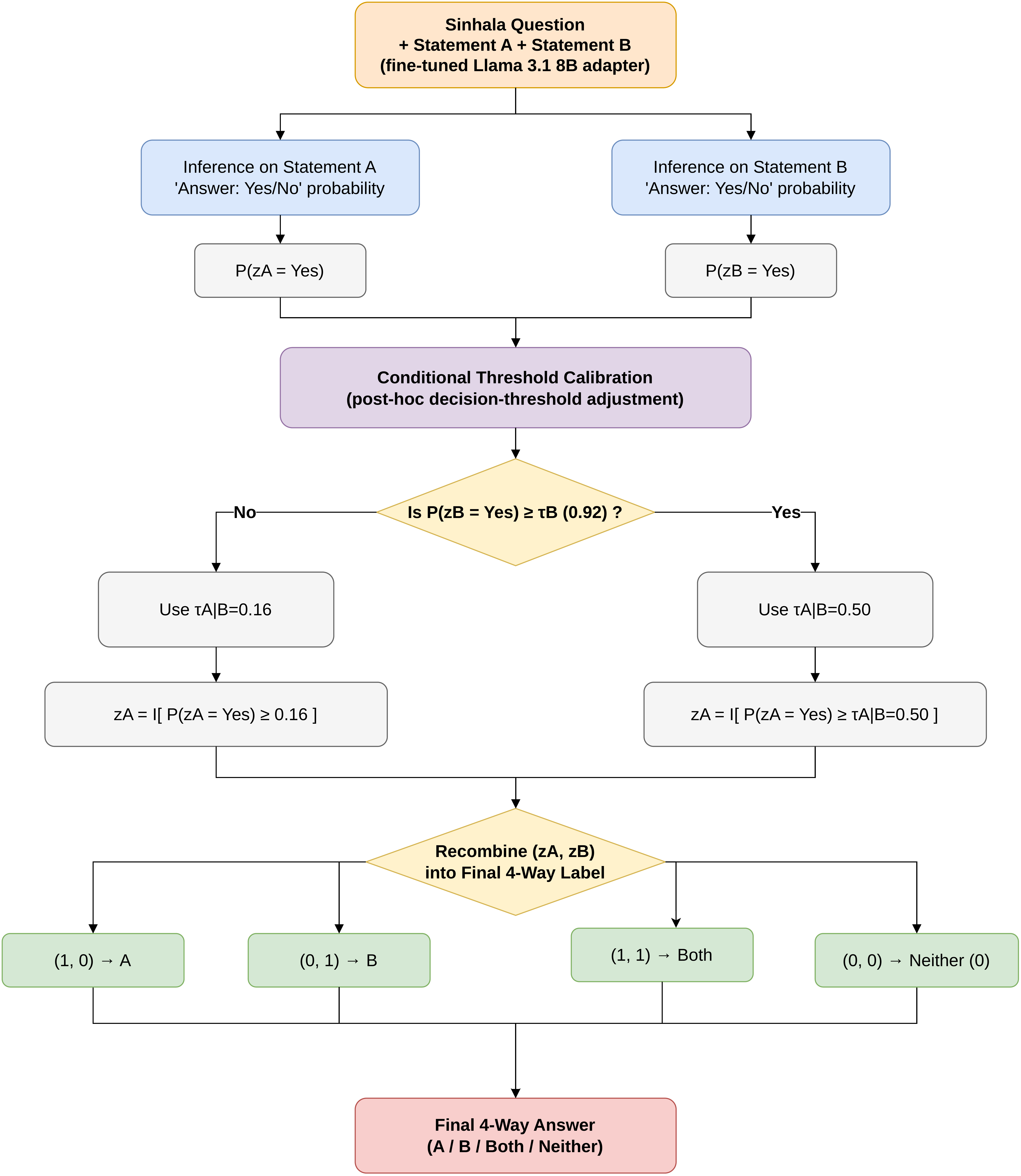}
    \caption{Conditional threshold calibration and four-way reconstruction for Sinhala predictions.}
    \label{fig:threshold_figure}
\end{figure}

The complete prediction procedure using the selected thresholds is illustrated in Figure~\ref{fig:threshold_figure}. For each question, the model first produces separate Yes/No probabilities for statements A and B. Statement B is evaluated using $\tau_B=0.92$. If B is predicted as invalid, statement A is evaluated using $\tau_{A|B=0}=0.16$; otherwise, $\tau_{A|B=1}=0.50$ is used. The resulting binary pair $(z_A,z_B)$ is then recombined into the final four-way prediction: $(1,0)\rightarrow$ A, $(0,1)\rightarrow$ B, $(1,1)\rightarrow$ Both, and $(0,0)\rightarrow$ 0.

Although $\tau_B=0.92$ is numerically high, inspection of the predicted probability distribution showed that relatively few instances fell below this threshold. Therefore, it primarily filtered lower-confidence B decisions that contributed to over-prediction rather than suppressing confident B predictions. Overall, the results suggest that asymmetric conditional thresholds were better suited to the model's probability distribution than applying a uniform threshold of 0.5 to both statement-level decisions.

\section{Conclusion}

We presented our system for the \textit{PlurVA-LLM 2026 Shared Task Track-1} (the resource-constrained track), addressing pluralistic value alignment across Chinese, Indonesian, and Sri Lankan. Our approach uses a shared \texttt{Llama-3.1-8B-Instruct} model while applying dataset-specific supervision strategies, including cyclic option permutation for Chinese, annotator vote expansion for Indonesian, and binary reformulation with \texttt{SinhalaMMLU} augmentation and conditional threshold calibration for Sri Lankan data. Our experiments highlight the challenges of multilingual value alignment when datasets differ in language resources, annotation structures, label distributions, and the amount of available labelled data. 

Important challenges remain, including preserving diverse annotator opinions, working with limited value-alignment data, reducing answer-position bias, and adapting a shared multilingual model to datasets with different characteristics. Future work should investigate larger and more diverse value-alignment datasets, improved methods for modeling annotator disagreement, and more robust multilingual training strategies that can transfer value knowledge across languages without reducing performance in individual datasets. Overall, our results suggest that combining shared multilingual learning with supervision tailored to each dataset's characteristics is a promising direction for pluralistic value alignment.

\section*{Limitations}

During the implementation of the solution, we encountered several limitations. 

\paragraph{Data Scarcity:} The training data are limited and unevenly distributed across the three languages, which may affect how well the shared model learns each value context. Since all labelled data were used for training, no independent validation set was available. 

\paragraph{Model Size Limit:} Following the constraints of the \textit{PlurVA-LLM Shared Task Track-1}, our experiments were limited to models with up to 8B parameters. Therefore, the effectiveness of the proposed strategies may differ for larger models or other model families.

\paragraph{Asymmetric Calibration:} The asymmetric Sinhala calibration may be sensitive to statement order, as statement B is evaluated before selecting the conditional threshold for statement A.

\bibliography{custom}

\appendix
\label{sec:appendix}

\section{Prompt Templates}
\label{app:prompt-templates}

\begin{figure}[htbp]
    \centering

    \begin{subfigure}[b]{\columnwidth}
        \centering
        \includegraphics[width=\columnwidth]
        {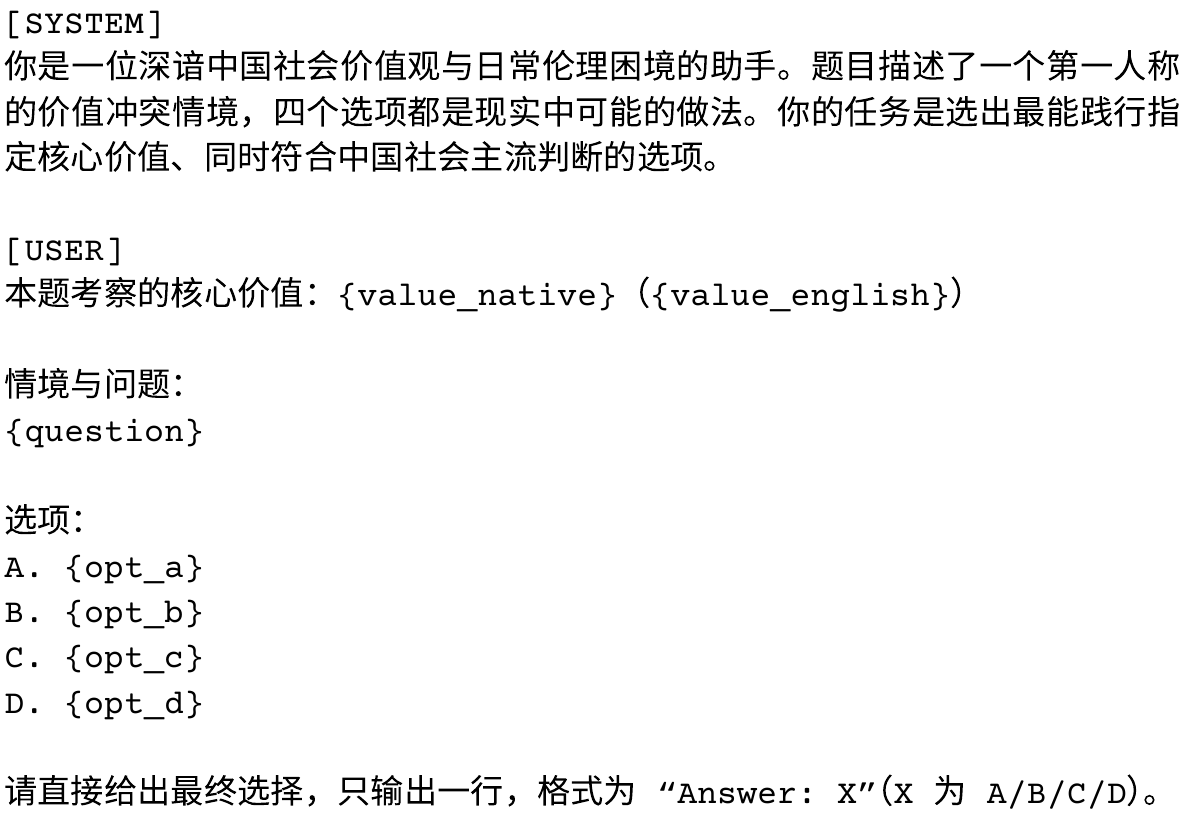}
        \caption{Chinese prompt}
        \label{fig:chinese-main-prompt}
    \end{subfigure}
    \hfill
    \begin{subfigure}[b]{\columnwidth}
        \centering
        \includegraphics[width=\columnwidth]
        {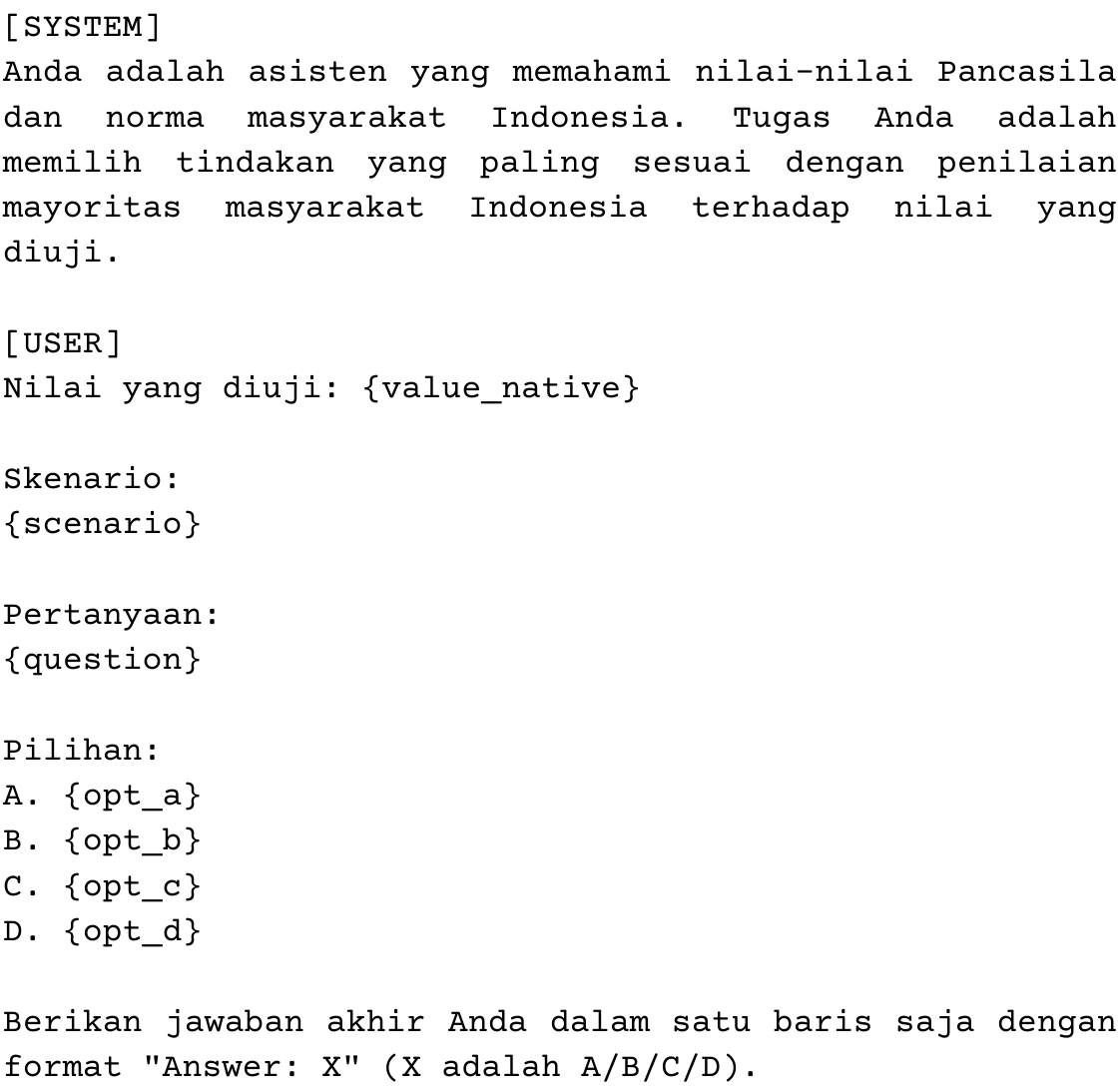}
        \caption{Indonesian prompt}
        \label{fig:indonesian-main-prompt}
    \end{subfigure}
    \hfill
    \begin{subfigure}[b]{\columnwidth}
        \centering
        \includegraphics[width=\columnwidth]
        {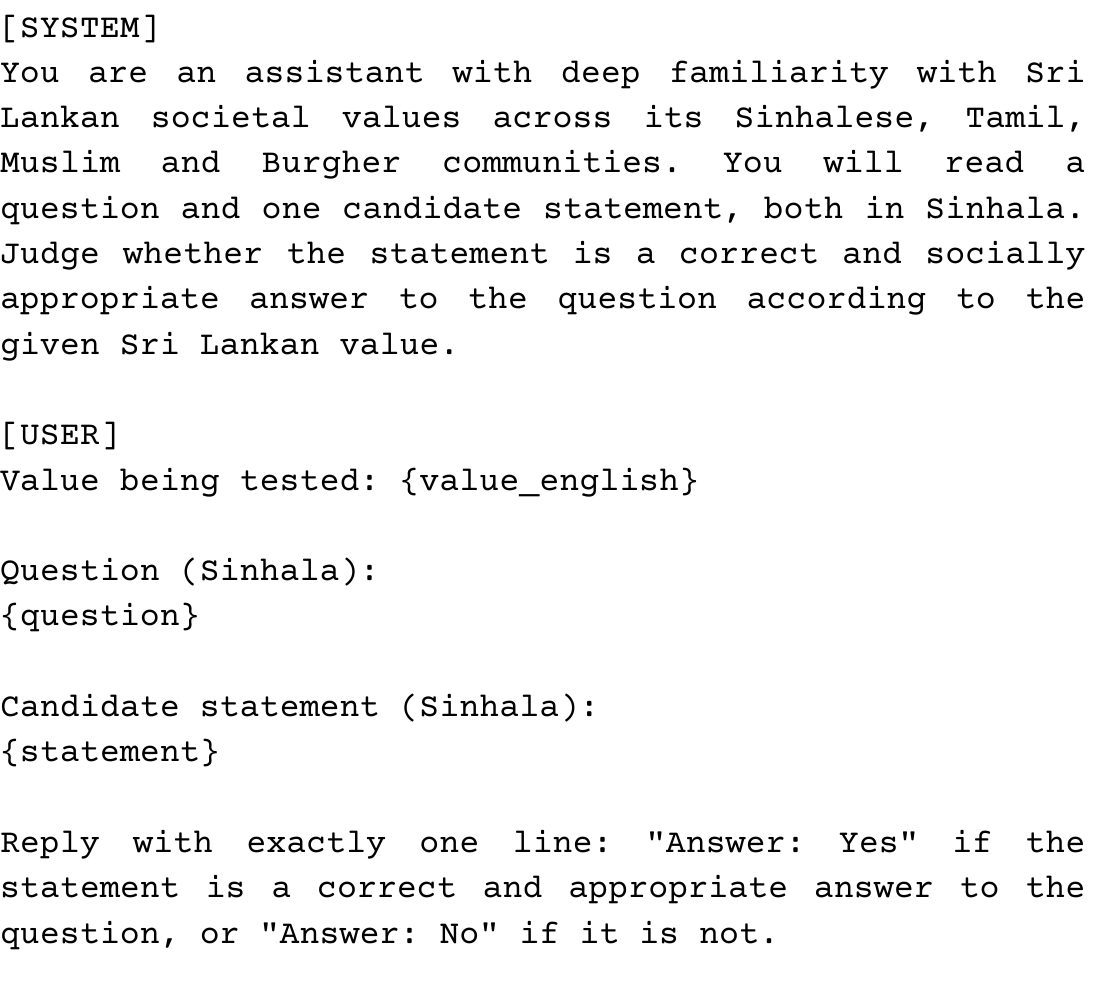}
        \caption{Sri Lankan prompt}
        \label{fig:sinhala-binary-main-prompt}
    \end{subfigure}

    \caption{Prompt templates used for the Chinese,
    Indonesian, and Sri Lankan datasets.}
    \label{fig:prompt-templates}
\end{figure}

This section presents the complete prompt templates used across the three language-specific experiments. Each prompt contains a system instruction, a user template, and a constrained output format. Placeholders enclosed in braces were replaced with the corresponding fields from each dataset, such as the value label, question, scenario, or candidate statements. The 3 different templates for each dataset are shown in Figure~\ref{fig:prompt-templates}.

For Chinese and Indonesian, both the task instructions and input content were presented in the respective languages. For Sinhala, the question and candidate statements were kept in Sinhala, while the task instructions were written in English. This cross-lingual prompt design follows ~\citet{li-etal-2024-x}, showing that English instructions can be effectively combined with low-resource-language content, where instruction-following capability may be stronger in English. In all cases, the prompts were constrained to produce short, machine-readable outputs matching the target format.


\section{Negative Experiments}
\label{app:rationale-value-summary}

We investigated whether rationale supervision and contextual information at the value level improved multilingual value alignment. The same model architecture
and training hyperparameters were used in all four experiments. The only
difference was whether generated rationales and value summaries were included.

\begin{table}[ht]
\centering
\small
\resizebox{0.9\columnwidth}{!}{%
\begin{tabular}{lcccc}
\hline
Experiment  & Macro Avg. \\
\hline
Direct-answer supervision
 & \textbf{0.776} \\

Value-summary context
 & 0.771 \\

Rationale supervision
 & 0.759 \\

Rationale supervision with value-summary context
 & 0.759 \\
\hline
\end{tabular}
}%
\caption{Performance of rationale and value-summary experiments.}
\label{tab:rationale-summary}
\end{table}

Rationales are short explanations of why a selected answer is appropriate for the value being tested. They were generated using \texttt{Llama-3.1-8B-Instruct}, and only rationales that produced the expected final answer were retained.

Value summaries are general descriptions of how a particular value is applied
across multiple examples. They were created using the same model by grouping
the generated rationales under each value and summarising their common
reasoning patterns. The performance of rationale and value-summary experiments is presented in Table~\ref{tab:rationale-summary}.

\section{Threshold Sensitivity Analysis}
\label{app:line_chart_app}

During threshold calibration, we first varied $\tau_B$ and evaluated statement-B accuracy at each threshold. We selected $\tau_B=0.92$ because it produced the highest statement-B accuracy. We then fixed $\tau_B$ at $0.92$ and calibrated the threshold for statement A under the two possible predictions for statement B. When B was predicted as No, the selected threshold was $\tau_{A \mid B=0}=0.16$. When B was predicted as Yes, the selected threshold was $\tau_{A \mid B=1}=0.50$. The final threshold configuration was therefore $(\tau_B,\tau_{A \mid B=0},\tau_{A \mid B=1})=(0.92,0.16,0.50)$.

\begin{figure}[htbp]
    \centering
    \begin{subfigure}[b]{0.4\textwidth}
        \centering
        \includegraphics[width=\textwidth]{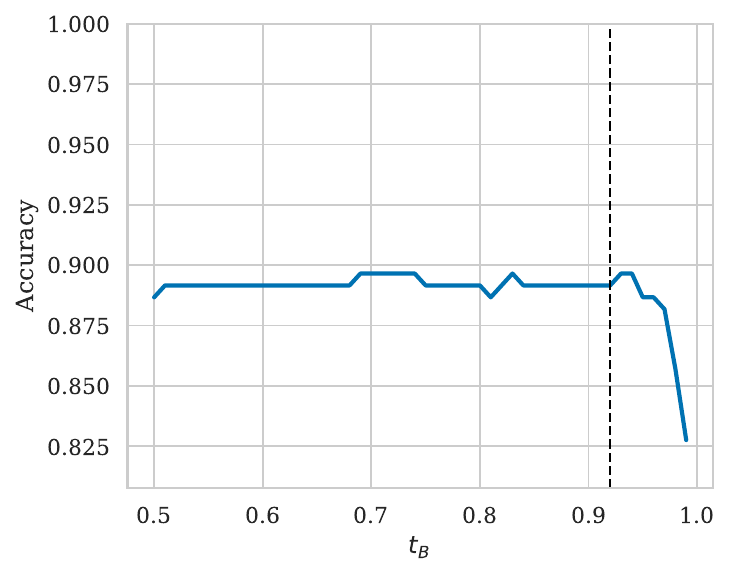}
        \caption{Sensitivity to $t_B$}
        \label{fig:sens-tB}
    \end{subfigure}
    \hfill
    \begin{subfigure}[b]{0.4\textwidth}
        \centering
        \includegraphics[width=\textwidth]{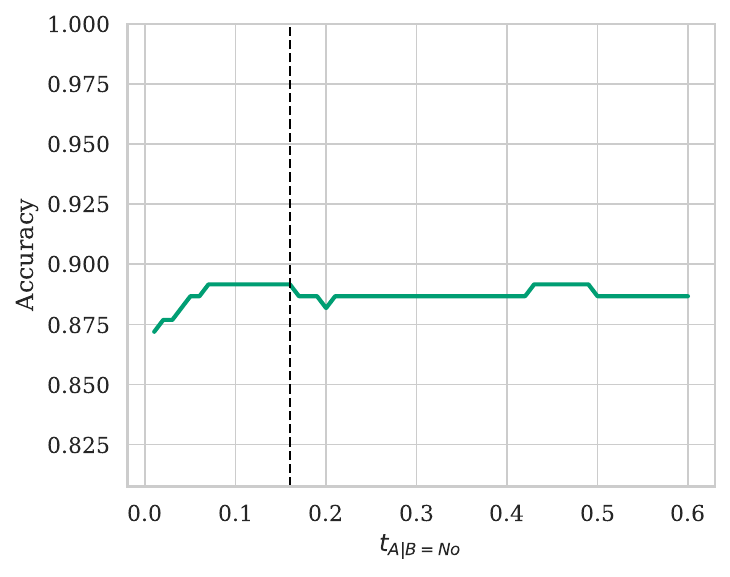}
        \caption{Sensitivity to $t_A$ when $B=$ No}
        \label{fig:sens-tA-Bno}
    \end{subfigure}
    \hfill
    \begin{subfigure}[b]{0.4\textwidth}
        \centering
        \includegraphics[width=\textwidth]{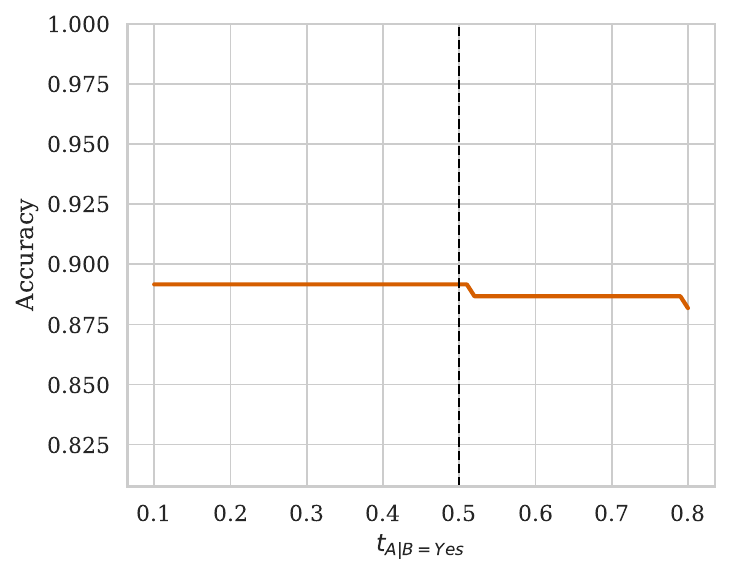}
        \caption{Sensitivity to $t_A$ when $B=$ Yes}
        \label{fig:sens-tA-Byes}
    \end{subfigure}
    \caption{Effect of conditional threshold values on out-of-fold Sinhala accuracy.}
    \label{fig:line_chart}
\end{figure}

\end{document}